\documentclass{article}
\usepackage{ijcai21}

\usepackage{times}

\usepackage{soul}
\usepackage{url}
\usepackage[hidelinks]{hyperref}
\usepackage[utf8]{inputenc}
\usepackage[small]{caption}
\usepackage{graphicx}
\usepackage{amsmath}
\usepackage{booktabs}
\usepackage[dvipsnames]{xcolor}

\usepackage{graphicx}
\usepackage{amsmath}
\usepackage{esvect}
\usepackage{outlines}

\usepackage[disable]{todonotes}

\usepackage[ruled,vlined]{algorithm2e}
\usepackage{anyfontsize}
\usepackage{t1enc}

\usepackage{enumitem}
\usepackage{multicol}
\usepackage{booktabs}
\usepackage{threeparttable}
\usepackage{multirow}
\usepackage{tikz}

\usepackage{pgfplots, pgfplotstable}
\usepgfplotslibrary{colorbrewer}
\pgfplotsset{compat=1.10}
\pgfplotsset{cycle list/Set1-6}

\usepackage{diagbox}

\usepackage{pgfplotstable} 

\newtheorem{example}{Example}

\newcommand\myeq{\mkern1.5mu{=}\mkern1.5mu}

\title{xRAI: Explainable Representations through AI}

\author{
Christian Bartelt$^1$\footnote{Contact Author}\and
Sascha Marton$^2$\And
Heiner Stuckenschmidt$^2$\\
\affiliations
$^1$Clausthal University of Technology, 38678 Clausthal-Zellerfeld\\
$^2$University of Mannheim, 68131 Mannheim\\
\emails
christian.bartelt@tu-clausthal.de,
marton@es.uni-mannheim.de,
heiner@informatik.uni-mannheim.de
}

\begin{document}

\maketitle
\begin{abstract}
We present xRAI an approach for extracting symbolic representations of the mathematical functions a neural network was supposed to learn from the trained network. The approach is based on the idea of training a so-called interpretation network that receives the weights and biases of the trained network as input and outputs the numerical representation of the function the network was supposed to learn that can be directly translated into a symbolic representation. We show that interpretation nets for different classes of functions can be trained on synthetic data offline using Boolean functions and low-order polynomials as examples. We show that the training is rather efficient and the quality of the results are promising. Our work aims to provide a contribution to the problem of better understanding neural decision making by making the target function explicit.  
\end{abstract}

\section{Introduction}

The ability of artificial neural networks to act as general function approximators has led to impressive results in many application areas. Limitations of early approaches have been overcome by using increasingly sophisticated architectures that make them almost universally applicable if enough data is available. The price for this universal applicability is the limited interpretability of the trained model. Overcoming this limitation is a subject of active research in the machine learning community \cite{guidotti2018survey}. Popular approaches for explaining the results of neural nets such as LIME \cite{deeplift_ribeiro2016should}, SHAP \cite{shap_lundberg2017unified} or LRP \cite{lrp_montavon2019layer} focus on the impact of the different attributes on the predictions of the model for certain examples. While this provides a partial explanation for individual examples, it does not really shed a light on the target function of the neural net training. 

In this paper, we investigate whether this function can be extracted post-hoc from a trained neural network. We assume that a neural network, which we call $\lambda$-Net in the following, has been trained to approximate a mathematical function from a certain function family. In this paper, we use Boolean functions or low arity and low-order polynomials as examples. However xRAI can be applied to any function family efficiently learnable by a neural network. For the case of low-order polynomials, this has been shown by \cite{andoni2014learning}. 

For each family of functions, we train a neural network called interpretation network ($\mathcal{I}$-Net). The $\mathcal{I}$-Net receives the weights and biases of a $\lambda$-Net as input and determines an approximation of a target function of the trained $\lambda$-Net. We train the $\mathcal{I}$-Net offline by systematically training $\lambda$-Nets on different functions from the family and using these trained networks as training examples for the $\mathcal{I}$-Net. 

The contribution of this paper are the following:  

\begin{itemize}
\item We propose interpretation networks ($\mathcal{I}$-Nets) as a means for the post-hoc interpretation of neural networks by extracting a mathematical functions from the trained network.
\item We present $\mathcal{I}$-Net designs for Boolean functions and low order polynomials by proposing corresponding encodings and loss functions.
\item We show that the proposed $\mathcal{I}$-Net designs can efficiently be trained offline and are capable of identifying the correct function with high level of accuracy.
\end{itemize}

The paper is structured as follows. We first briefly present the idea of $\mathcal{I}$-Nets as a tool for analyzing trained neural networks, we then present $\mathcal{I}$-Net designs and training procedures for Boolean Functions and low order polynomials. Finally, we experimentally evaluate the approach with respect to its prediction quality as well as the impact of amount of training data and learning epochs before concluding with a discussion of the approach.  

\section{The xRAI Approach}

\begin{figure}[ht]
    \centering
    \includegraphics[width=0.9\columnwidth]{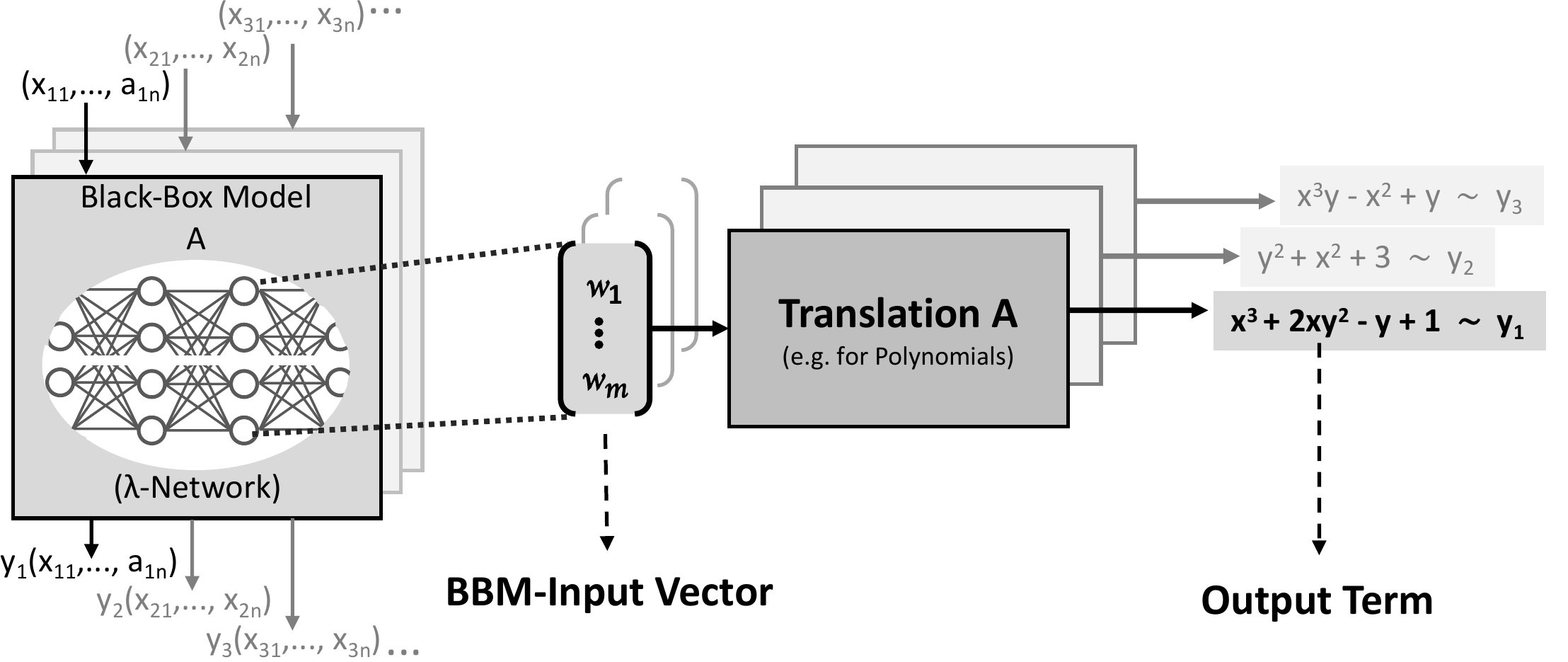}
    \caption{Overview of the Approach}
    \label{fig:overview}
\end{figure}

To overcome the introduced research challenge, we have developed a translation technique as depicted in Fig.~\ref{fig:overview}. Thereby a trained neural net (a so-called $\lambda$-Net) can be translated in a human understandable representation of its target function. For example, assuming the trained neural network (black box model) can be also represented as a mathematical function then the goal of the translation (resp. interpretation) is the determination of a semantically equivalent, algebraic term.  Following a white-box approach, the input term of the interpretation is derived by the $\lambda$-Net parameters and represented as a vector of the weights and biases (\underline{B}lack\underline{B}ox\underline{M}odel-Input Vector). The output of the translation is a numeric encoding of an algebraic expression. In Fig.~\ref{fig:overview} the output terms are polynomial expressions for example.

The key aspect of xRAI is the usage of neural networks to process the mentioned interpretation of other neural networks. Therefore we introduce so-called \textit{interpretation networks} ($\mathcal{I}$\textit{-Nets}) as a core element of xRAI. An $\mathcal{I}$-Net is neural network that is able to translate a $\lambda$-Net into a numeric output vector which encodes a mathematical term. The $\mathcal{I}$-Net has to be trained by a sufficient parameter set of $\lambda$-Nets (as inputs) and corresponding algebraic expressions (as outputs) which are compliant with their target functions. Hence we conceptualize this determination of a translation challenge as a machine learning (ML) problem. To formulate the ML problem we have to answer two questions in the following sections:
\begin{itemize}
    \item How can a mathematical expression be encoded suitably as a numeric output vector of a neural net?
    \item How can we train an $\mathcal{I}$-Net and especially what is a suitable loss function?
\end{itemize}

In the following two sections we will answer both questions more in detail.
\section{Algebraic Expressions Determined by Neural Networks}

\begin{figure}[ht]
    \centering
    \includegraphics[width=0.9\columnwidth]{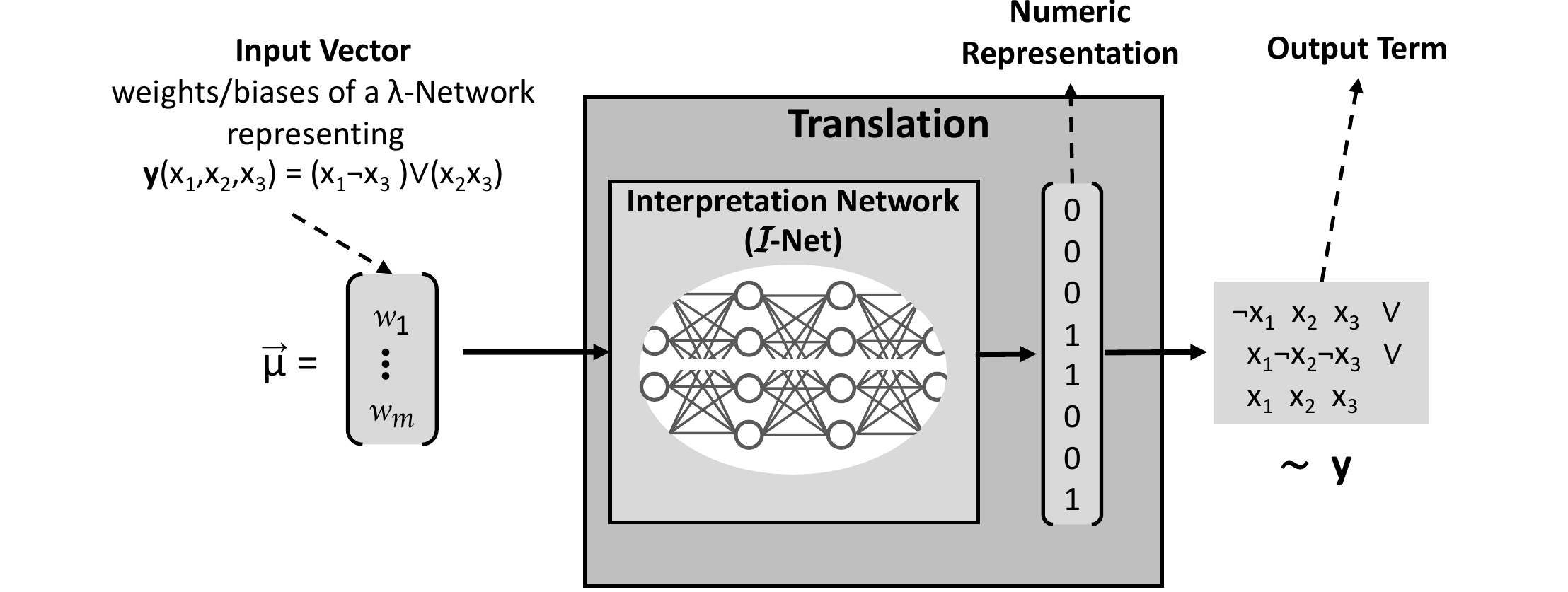}
    \caption{Boolean Function Encoding as $\mathcal{I}$-Net Output}
    \label{fig:BoolEncoding}
\end{figure}

To illustrate the translation of neural networks into algebraic expressions (cf. $\mathcal{I}$-Net in Fig.~\ref{fig:BoolEncoding}), we use exemplarily two function families. The first one is defined by the algebra of polynomials and the second one is defined by the Boolean algebra. Therefore the initial translation result has to be encoded as a numeric vector, because we use a neural network as the translation function. In a second part of the translation the numeric output vector can decode/represent as the corresponding term of an algebraic expression.

\subsection{Encoding of Boolean Terms as Numeric Vectors}
For a numeric representation of Boolean expressions, we use a certain encoding of Boolean terms in the canonical disjunctive normal form (CDNF). Each Boolean function can be represented as a unique expression of the CDNF. The number of possible minterms in CDNF resp. DNF expressions determined by the maximum of variables in the Boolean functions which should be represented and calculated by $2^n$. For example all Boolean functions with at most three variables can be represented as a logical disjunction of the following $2^3=8$ minterms. 

\begin{multicols}{2}
\begin{enumerate}[start=0, leftmargin=0.5cm]
{\fontsize{8}{6}\selectfont
    \item $MIN_{000} = \neg x_{1} \neg x_{2} \neg x_{3}$
    \item $MIN_{001} = \neg x_{1} \neg x_{2} x_{3}$
    \item $MIN_{010} = \neg x_{1} x_{2} \neg x_{3}$
    \item $MIN_{011} = \neg x_{1} x_{2} x_{3}$
    \item $MIN_{100} = x_{1} \neg x_{2} \neg x_{3}$
    \item $MIN_{101} = x_{1} \neg x_{2} x_{3}$
    \item $MIN_{110} = x_{1} x_{2} \neg x_{3}$
    \item $MIN_{111} = x_{1} x_{2} x_{3}$ 
}
\end{enumerate}
\end{multicols}

For the encoding of a Boolean function the upper mentioned enumeration schema of minterms is the key factor. Hence for the encoding of Boolean expressions with at most 3 variables we can use a vector with 8 scalars (one scalar for each minterm ordered by the upper enumeration schema).
\paragraph{Mapping between Boolean Expressions and Numeric Vectors}
We define that the CDNF representation of a Boolean expression contains a specific minterm if and only if the corresponding scalar in the $\mathcal{I}$-Net output vector is greater than $0.5$.
To illustrate this encoding vividly, we introduce the following example of a Boolean term (cf. Fig.~\ref{fig:BoolEncoding}):

\begin{example}[Boolean expression]
\[(x_1 \wedge \neg x_3) \vee (x_2 \wedge x_3)\]
\end{example}

This term can be transformed in a Boolean expression in CDNF as following:

\[\neg x_1x_2x_3 \vee x_1\neg x_2\neg x_3 \vee x_1x_2x_3\]

Now we can represent this CDNF-expression as a numeric vector: The expression contains the third, fourth, and seventh minterm of the our upper introduced minterm enumeration. Accordingly, the following vector represent the Boolean function $\beta(x_1, x_2, x_3,) = (x_1 \wedge \neg x_3) \vee (x_2 \wedge x_3)$:
\begin{equation}
    \vec{b} = (0, 0, 0, 1, 1, 0, 0, 1)^{\top}
\end{equation}

Using this encoding principle we can represent each numeric vector as a Boolean expression and vice versa. In general each Boolean function respectively CDNF-expression $\beta(x_1, \ldots, x_n)$ can be represented as a vector $\vec{b}$ as following:

{
\setlength{\interspacetitleruled}{-.4pt}%
\begin{algorithm}[ht]
  \eIf{$\forall b_{i}<0.5 $}{
   $\beta(x_1, \ldots, x_n) = 0$
   }{
   {\fontsize{7}{9}\selectfont $\vec{b} = (b_1  \ldots b_{2^{n}})^{\top} \Leftrightarrow \beta(x_1, \ldots, x_n) := \! \underset{ b_{i} \ge 0.5}{\bigvee} MIN_{(i-1)_{2}}$}
  }

\end{algorithm}
}

where $MIN_{(x)_{2}}$ is the $x.$ minterm of the enumeration above.

\subsection{Encoding of Polynomial Terms as Numeric Vectors}
As usually, we consider polynomials as expressions consisting of variables (indeterminates) and coefficients, that involves only the operations of addition, subtraction, multiplication, and non-negative integer exponents of variables. The degree of a polynomial is determined by the largest degree of its monomial terms with non-zero coefficient. The degree of a monomial term is the sum of the exponents of the contained variables.

\begin{example}[Polynomial Expression A]
\[x^3 - 2x^2 + 5\]
\end{example}
is a polynomial (with degree $3$) over only one variable $x$. 
\begin{example}[Polynomial Expression B]
\[x^2 + 3xyz - 2y + z \]
\end{example}
is a polynomial (with degree $3$) over $3$ variables ($x$, $y$, $z$).

\paragraph{Mapping between Polynomial Expressions and Numeric Vectors}
Each polynomial expression can be represented as a unique numeric vector of its coefficients. For univariate polynomials this encoding is straightforward. For example the cubic polynomial $f(x) = x^3 - 2x^2 + 5$ can be encoded as the vector of coefficients $\vec{p} = (1,-2,0,5)$. With increasing number of indeterminates of the polynomial, the number of possible monomial terms expands rapidly and is calculated by \(\binom{n+d}{d}\) where $d$ is the degree and $n$ the number of variables. Accordingly, a cubic, bivariate polynomial has \(\binom{2+3}{3} = 10\) possible coefficients each belonging to one of the following monomials

\begin{multicols}{2}
\begin{enumerate}[leftmargin=1.25cm, start=0]
    \item $x^3 y^0 = x^3$
    \item $x^0 y^3 = y^3$
    \item $x^2 y^1 = x^2y$
    \item $x^1 y^2 = xy^2$
    \item $x^2 y^0 = x^2$
    \item $x^0 y^2 = y^2$
    \item $x^1 y^1 = xy$
    \item $x^1 y^0 = x$
    \item $x^0 y^1 = y$
    \item $x^0 y^0 = 1$
\end{enumerate}
\end{multicols}

As an example, we could represent the bivariate polynomial of degree $3$ $f(x,y)=3x^3 - 5xy^2 + x^2 - 3y - 4$ as a vector of the coefficients 
\[\vec{p} = (3,0,0,-5,1,0,0,0,-3,-4)^{\top}\]
according to the encoding above. The order of scalars in the vector $\vec{p}$ is sorted by the upper depicted enumeration of monomials.

\section{Training of Interpretation Networks ($\mathcal{I}$-Nets)}

Before we can interprete neural nets, we need a performant translation function (the $\mathcal{I}$-Net) which is able to determine the corresponding mathematical expression very accurately. Therefore we have to train an $\mathcal{I}$-Net for each function family which should be interpreted as an explicit algebraic expression. The basis for the training of $\mathcal{I}$-Nets is a sufficient example set of functions which are represented two ways: Each training example consist of a set of neural network parameters (weights and biases of a $\lambda$-Net) together with the corresponding numeric representation of the mathematical expression. Two kinds of numeric representations of algebras were presented in the last section. The $\mathcal{I}$-Net training process starts with the generation of training examples and is described in the following four steps:

\subsection{Step 1: Function Selection for the Training of $\lambda$-Net Examples}
At the beginning of the training process, we need a set of functions $\lambda_{1,2,...,i} \in \Lambda$ to train a set of $\lambda$-Nets (i.e. neural nets that were trained to learn a target function $\lambda$). Therefore, we must select at first an algebra and design an appropriate numeric encoding of its function terms (cf. the examples in the previous section). Afterward, we can generate a random set of functions $\Lambda$ from the selected algebra resp. function family along with a synthetic set of training examples as
\begin{equation*}
    \left\{ \mathcal{D}_{\lambda_i} = \left\{(\mathbf{x_{ij}},  y_{ij})\right\}^{s}_{j=1}\right\}^{|\Lambda|}_{i=1} , 
\end{equation*}
where $\mathcal{D}_{\lambda_i}$ is the training set for a function $\lambda_i$,  $\mathbf{x_{ij}}$ a random data point of dimensionality $D$ and $y_{ij} = \lambda_i \left( x_{i,j,1}, x_{i,j,2}, \dots ,x_{i,j,D} \right)$ is the corresponding value.

\subsection{Step 2: Generation of $\lambda$-Nets as Training Inputs}
Next, we need to train a set of $\lambda$-Nets. Thereby, we need to train one model for each function $\lambda_{1,2,...,i} \in \Lambda$ based on the previously generated training sets $\mathcal{D}_{\lambda_i}$. As a result, we have a set of trained neural networks, each approximating a function $\lambda_i$. At this point, we know the target function $\lambda_i$ for each $\lambda_i$-Model.

\subsection{Step 3: Example Sets for the Training of an $\mathcal{I}$-Net}
In the third step, we need to generate a training set for our $\mathcal{I}$-Net based on the neural network models of the functions $\lambda_{1,2,...,i} \in \Lambda$ ($\lambda$-Nets) in our example set trained in the previous step. The objective is finding the function, the neural network was supposed to learn. Since we know this particular function ($\lambda_i$) for the $\lambda_i$-Models trained on the synthetically created datasets $\mathcal{D}_{\lambda_i}$, we can generate our training set for the $\mathcal{I}$-Net $\mathcal{D}_{\mathcal{I}_a}$ as follows:
\begin{equation*}
    \mathcal{D}_{\mathcal{I}_a} = \left\{ (\boldsymbol{\mu_i}, \lambda_i  ) \right\}^{|\Lambda|}_{i=1},
\end{equation*}
where $\boldsymbol{\mu_i}$ are the learnt model parameters (weights and biases of the $\lambda_i$-Model) and $\lambda_i$ is a numeric representation of corresponding function (cf. Fig.~\ref{fig:BoolEncoding} and Equation 1 and 2).

Depended on the algebra of target functions a suitable loss function has to be used during the training process of the $\mathcal{I}$-Net. To explain this task we describe two variants of loss functions for our both introduced algebras: 

\subsubsection{Loss Function for the Training of a Boolean $\mathcal{I}$-Net}
For the training of our $\mathcal{I}$-Net on Boolean algebra, we can straight forward calculate the loss function as a cross-entropy loss for multi-label classification by summing up the binary cross-entropy loss for each sample as
\begin{equation*}
    \mathcal{L}_{Boolean} = \frac{1}{s \times N} \sum^{s}_{i=1} \sum^{N}_{j=1} | \lambda_{i, j} - \hat{\lambda}_{i, j} | ,
\end{equation*}
where $s$ is the number of evaluations and \(N=2^n\) is the number of possible minterms, \(\lambda_{i, j}\) states whether the minterm at index $j$ is present in the CDNF of the $i$-th real and \(\hat{\lambda}_{i, j}\) whether the minterm at index $j$ is present in the CDNF of the $i$-th predicted function.

\subsubsection{Loss Function for the Training of a Polynomial $\mathcal{I}$-Net}
While the loss function is straight forward for the Boolean algebra, it becomes more difficult for polynomials since there is no direct relationship between the monomials of a polynomial and the function value. One intuitive loss function is the integral difference between target function and interpreted function in a specific subspace. But the determination of integrals of multivariate polynomials is a hard to compute \cite{fu_multivariate_2012} and cannot be applied reasonably in our machine learning loop. Maybe it can be mitigated with help of advanced techniques \cite{barvinok_integration_2007} and is matter of further research. For the application of xRAI we have used a simplified and more efficiently computable loss function based on the difference of function values.  Therefore we need to select a set of $m$ random data points and compare the function values of the target polynomial \(\lambda_i(\mathbf{x_j})\) and the interpreted polynomial \(\hat{\lambda}_i(\mathbf{x_j})\) based on a sample. In this case, $\mathbf{x_j}$ is a vector of variable values with length $n$.  We can do this as follows:
\begin{equation*}
    \mathcal{L}_{Polynomial} = \frac{1}{s \times m} \sum^{s}_{i=1} \sum^{m}_{j=1} | \lambda_i(\mathbf{x_j}) - \hat{\lambda}_i(\mathbf{x_j}) | ,
\end{equation*}
where $s$ is the number of evaluations.

\subsection{Step 4: Training of an Interpretation Net}
Now we are able to train an $\mathcal{I}$-Net. As a result, the $\mathcal{I}$-Net conforms to an approximation of a semantic mapping from the internals of a neural network to expression of the target function of the neural network.

\section{Post-hoc Interpretation of Neural Networks by Applying  Interpretation Nets}

Having trained an appropriate $\mathcal{I}$-Net, we are now able to interpret neural nets ($\lambda$-Nets) explicitly as expressions of a selected algebra. Therefore, we just need to pass the weights and biases of the $\lambda$-Net to the $\mathcal{I}$-Net and make a $\lambda$-Net-Interpretation (cf. Fig.~\ref{fig:overview}, \ref{fig:BoolEncoding}). The quality of this interpretation is depend on the selection of an appropriate pre-trained $\mathcal{I}$-Net. This $\mathcal{I}$-Net must have been trained for the right function family resp. algebra. For example, if the target function, which is approximated by the $\lambda$-Net, is a cubic polynomial with $4$ variables then we need an $\mathcal{I}$-Net which was trained for polynomial expressions with at least degree $3$ and $4$ variables. In relation to other interpretability approaches based on surrogate models, the $\mathcal{I}$-Net of xRAI is able to translate/interpret a neural net immediately. Consequently the learning of a surrogate model with the $\lambda$-Net results can be omitted in many cases. In the following section we present the evaluation results for two function families - Boolean functions and polynomials.

\section{Evaluation}
We evaluated xRAI empirically based on an experiment series of post-hoc interpretations of feedforward artificial neural networks. Each neural network to interpret in our experiments has learned a ML model representing any mathematical function of a certain function family within the Boolean algebra or the algebra of polynomials. For this purpose, we conducted three experiment sub-series: 
The first experiment sub-series evaluates the $\mathcal{I}$-Net performance for both algebras considering different numbers of variables and therefore complexities. Thereby, we want to empirically prove in general that it is possible to learn a translation from the model internals of an neural network to semantically well-interpretable formal expressions and we are therefore able to extract the target function from an already trained neural network. 
Within the second experiment series, we compare the $\mathcal{I}$-Net performance with the $\lambda$-Net performance at different $\lambda$-Net training levels. This allows us to show that xRAI is able to interpret neural networks very accurately at arbitrary stages of the $\lambda$-Net training process and puts the results from the first experiment series into context. 
The third experiment series evaluates the $\mathcal{I}$-Net performance for different sample sizes of the training set. Thereby, we want to address the scalability of xRAI and identify if and how the performance of the $\mathcal{I}$-Net can be increased once a larger set of $\lambda$-Nets is used for the training.
In the following, we will first summarize the experimental setup, followed by the results of the three experiment series introduced above.

\subsection{Experimental Setup}
 The relevant parameters used for the training of the $\lambda$- and $\mathcal{I}$-Nets are summarized in Table~\ref{tab:net-params}. For the $\lambda$-Nets we selected the parameters based on the findings of \cite{andoni2014learning} which proved that neural networks with a single hidden layer are able to learn polynomials efficiently using SGD with $5 \times sparsity$ neurons, where in our case \(sparsity = \binom{k+n}{n}\). For comparability reasons, we selected those parameters similarly for the Boolean algebra. As activation function, we selected ReLU which is defined as $\text{max}(0, x)$ due to their good performance, especially in supervised settings and computational efficiency~\cite{glorot2011deep}. The remainder of the parameters of the $\mathcal{I}$-Nets were selected based on coarse to fine tuning to optimize the performance.

\begin{table}[ht]
\centering
\resizebox{\columnwidth}{!}{
\begin{threeparttable}
\begin{tabular}{@{}c|cccc@{}}
\toprule
\multirow{2}{*}{\textbf{Parameter}} & \multicolumn{2}{c}{\textbf{{\boldmath$\lambda$}-Hyperparameters }} & \multicolumn{2}{c}{\textbf{{\boldmath$\mathcal{I}$}-Hyperparameters }} \\ 
& Boolean & Polynomial & Boolean & Polynomial \\ \midrule
Network Structure\tnote{a} & \([5 \times 2^n]\) & \([5 \times \binom{n+d}{d}]\) & $[2048]$ & $[2048]$  \\
Hidden Layer Activation & ReLU & ReLU & ReLU & ReLU \\
Batch Size &  $2^n$ & 64 & 64 & 128 \\ %
Optimizer & SGD & SGD & \begin{tabular}[c]{@{}c@{}}Adadelta\end{tabular} & \begin{tabular}[c]{@{}c@{}}Adam\end{tabular} \\
Loss Function & \begin{tabular}[c]{@{}c@{}}Binary \\ Crossentropy\end{tabular} & MAE & $\mathcal{L}_{Boolean}$ & $\mathcal{L}_{Polynomial}$ \\
Training Epochs & 200 & 200 & 100 & 100\\ 
\bottomrule
\end{tabular}
\begin{tablenotes}
\item[a] \footnotesize In this context, $n$ stands for the number of variables and $k$ for the degree of a polynomial.
\item[b] \footnotesize See \cite{kingma2014adam} for details.
\item[c] \footnotesize See \cite{zeiler2012adadelta} for details.
\end{tablenotes}
\end{threeparttable}
}
\caption[Parameters for $\lambda$- and $\mathcal{I}$-Net]{Parameters for $\lambda$- and $\mathcal{I}$-Net}
\label{tab:net-params}
\end{table}

For the evaluation, we generated three datasets for each algebra, differing in the complexity based on the number of variables, which we use for the training of $\mathcal{I}$-Nets. As usual, we randomly split the available data into disjoint training, validation and test sets\footnote{One exception is the training of $\lambda$-Nets for the Boolean algebra since we need to use the complete dataset (which has the size $2^n$) to ensure that the Boolean functions are defined explicitly.}. The settings used for the training of the $\lambda$-Nets and the generation of the dataset for the $\mathcal{I}$-Net are summarized in Table~\ref{tab:dataset-specifications}.

\begin{table}[ht]
\centering
\resizebox{0.9\columnwidth}{!}{
\begin{threeparttable}
\begin{tabular}{@{}c|cc@{}}
\toprule
\textbf{Parameter} & \textbf{Boolean Algebra} & \textbf{Polynomials} \\  \midrule
Number of Variables ($n$) & ${4, 5, 6}$ & ${4, 5, 6}$ \\
Degree ($d$) & $0$ & $3$ \\
$n$-Range\tnote{a} & $\{0, 1\}$ & $[-1, 1]$\\
Coefficient Range\tnote{b} & $\{0, 1\}$ & $[-10, 10]$ \\ \midrule
\begin{tabular}[c]{@{}c@{}}$\lambda$-Net Training Set Size\end{tabular} & $2^n$ & $1,000$ \\
\begin{tabular}[c]{@{}c@{}}$\mathcal{I}$-Net Training Set Size\end{tabular} & $65,536$ & $50,000$ \\
\bottomrule
\end{tabular}
\begin{tablenotes}
\item[a] \footnotesize The \textit{$n$-Range} determines the range of the variable values used for the generation of the datasets.
\item[b] \footnotesize  The \textit{Coefficient Range} determines the range of the coefficients used for the random generation of the functions.
\end{tablenotes}
\end{threeparttable}
}
\caption[Dataset Specifications]{Dataset Specifications}
\label{tab:dataset-specifications}
\end{table}
Since the training of $\geq 50,000$ neural networks is extremely time-consuming, all experiments were conducted without repetition using a fixed random seed to ensure the reproducibility of our results and the comparability among $\lambda$-Nets for different functions.

\subsection{Experimental Results}

\subsubsection{Experiment Series 1: $\mathcal{I}$-Net Performance for Different Algebras and Complexities}

Within the first experiment, we want to evaluate the performance of the $\mathcal{I}$-Net based on the Boolean algebra and polynomials for different complexities in terms of the number of variables (cf.~Table~\ref{tab:inet-performance}).

\begin{table}[ht]
\centering
\resizebox{0.9\columnwidth}{!}{
\begin{threeparttable}
\begin{tabular}{cc|ccc}
\toprule
\multicolumn{2}{c}{\textbf{\textbf{\diagbox{Algebra (Score)}{Variables}}}} & \textbf{n=4} & \textbf{n=5} & \textbf{n=6} \\ \midrule
\multirow{2}{*}{\textbf{\begin{tabular}[c]{@{}c@{}}Boolean Algebra\\ (Accuracy)\end{tabular}}} & $\mathcal{I}$-Net & 1.00 & 1.00 & 0.88 \\ 
 & Na\"ive Baseline\tnote{a} & 0.50 & 0.50 & 0.50 \\ \midrule
\multirow{2}{*}{\textbf{\begin{tabular}[c]{@{}c@{}}Polynomials\\ (MAE)\end{tabular}}} & $\mathcal{I}$-Net & 2.43 & 3.47 & 4.91 \\ 
 & Na\"ive Baseline\tnote{b} & 14.94 & 16.55  & 18.97 \\ \bottomrule
\end{tabular}
\begin{tablenotes}
\item[a] \footnotesize The na\"ive baseline for the Boolean algebra equals a random guess whether each minterm is contained in the predicted DNF.
\item[b] \footnotesize  For the polynomials we can estimate the na\"ive baseline by selecting a number of samples (in our case $12,500$ which equals the size of the test set), randomly guessing a value for each coefficient based on the ranges defined in Table~\ref{tab:dataset-specifications} and evaluate the resulting polynomial similar to $\mathcal{L}_{Polynomial}$.
\end{tablenotes}
\end{threeparttable}
}
\caption[$\mathcal{I}$-Net Performance Overview]{$\mathcal{I}$-Net Performance Overview}
\label{tab:inet-performance}
\end{table}

For the Boolean algebra, the $\mathcal{I}$-Net was able to predict the Boolean function perfectly on all evaluated samples for $n\myeq4$ and $n\myeq5$. For $n\myeq6$ we still achieved very accurate results with $0.88$, which is significantly higher than a na\"ive baseline. However, we can observe that for an increasing complexity, the overall accuracy decreases when the remainder of the settings of the training process (e.g.~number of $\mathcal{I}$-Net training epochs and training data sample size) remain fixed.

For polynomials, we can similarly observe that the error increases along with the complexity of the polynomials and therefore the learning problem. However, once we compare the results of the $\mathcal{I}$-Net with a na\"ive baseline we can observe a significant performance increase for all complexities.
Accordingly, we are able to prove empirically that it is possible to extract the target function solely from the weights and biases of a neural network very accurately for two example algebras and at different levels of complexity.

\subsubsection{Experiment Series 2: $\mathcal{I}$-Net Performance Comparison for $\lambda$-Nets with Different Training Levels}

\begin{figure}[ht]
    \begin{tikzpicture}[scale=0.90]
    \begin{axis}[
          xlabel={$\lambda$-Net Training Epochs},
          ylabel style={align=center},
          ylabel={Accuracy},
          width=1\columnwidth,
          height=0.6\columnwidth,
          line width=1.0pt,
          mark size=2.5pt,
          every axis/.append style={font=\normalsize},
          ymax=1.05,
          ymin=0.35,
          ytick={0.4, 0.6, 0.8, 1},
          xmax = 200,
          xmin = 0,
          label style={font=\scriptsize},
          tick label style={},
          legend columns=3,
          legend style={font=\tiny, cells={align=center}, at={(0.5, -0.3)}, anchor=north, line width=0.4pt},
          every node near coord/.style=above,
          ]
      
      \addplot +[color=MidnightBlue, style=dashed] table[x expr=\thisrowno{0} , y=n4_lambda, col sep=semicolon] {./data/multi_epoch_analysis_results_boolean.txt}; 
      \addlegendentry{Average $\lambda$-Net $n\myeq4$}
      
      \addplot[color=Dandelion, style=dashed] table[x expr=\thisrowno{0} , y=n5_lambda, col sep=semicolon] {./data/multi_epoch_analysis_results_boolean.txt}; 
      \addlegendentry{Average $\lambda$-Net $n\myeq5$}
      
      \addplot[color=PineGreen, style=dashed] table[x expr=\thisrowno{0} , y=n6_lambda, col sep=semicolon] {./data/multi_epoch_analysis_results_boolean.txt}; 
      \addlegendentry{Average $\lambda$-Net $n\myeq6$}
      
      \addplot +[color=MidnightBlue] table[x expr=\thisrowno{0} , y=n4_int, col sep=semicolon] {./data/multi_epoch_analysis_results_boolean.txt}; 
      \addlegendentry{$\mathcal{I}$-Net $n\myeq4$}
      
      \addplot[color=Dandelion] table[x expr=\thisrowno{0} , y=n5_int, col sep=semicolon] {./data/multi_epoch_analysis_results_boolean.txt}; 
      \addlegendentry{$\mathcal{I}$-Net $n\myeq5$}      
      
      \addplot[color=PineGreen] table[x expr=\thisrowno{0} , y=n6_int, col sep=semicolon] {./data/multi_epoch_analysis_results_boolean.txt}; 
      \addlegendentry{$\mathcal{I}$-Net $n\myeq6$}      
    \end{axis}

    \end{tikzpicture}
    \caption[$\lambda$ and $\mathcal{I}$-Net Accuracy Comparison for different $\lambda$-Net Training Epochs on Boolean Algebra]{$\lambda$ and $\mathcal{I}$-Net Accuracy Comparison for different $\lambda$-Net Training Epochs on Boolean Algebra}
    \label{fig:comparison_lambda_inet_boolean}

\end{figure}
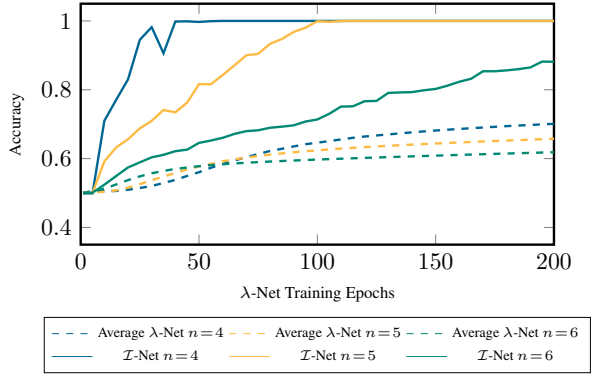

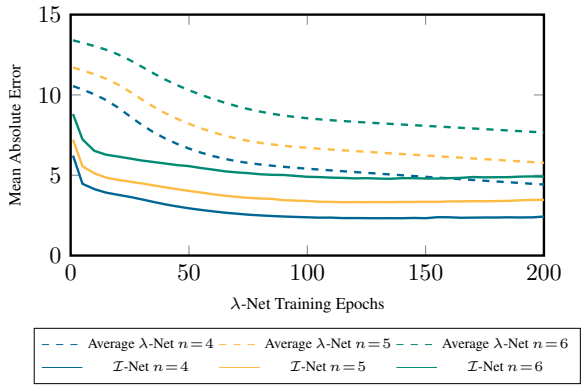
\begin{figure}[ht]
    \begin{tikzpicture}[scale=0.90]
    \begin{axis}[
          xlabel={$\lambda$-Net Training Epochs},
          ylabel style={align=center},
          ylabel={Mean Absolute Error},
          width=1\columnwidth,
          height=0.6\columnwidth,
          line width=1.0pt,
          mark size=2.5pt,
          every axis/.append style={font=\normalsize},
          ymax=15,
          ymin=0.0,
          ytick={15, 10, 5, 0},
          xmax = 200,
          xmin = 0,
          label style={font=\scriptsize},
          tick label style={},
          legend columns=3,
          legend style={font=\tiny, cells={align=center}, at={(0.5, -0.3)}, anchor=north, line width=0.4pt},
          every node near coord/.style=above,
          ]
      
      \addplot +[color=MidnightBlue, style=dashed] table[x expr=\thisrowno{0} , y=n4_lambda, col sep=semicolon] {./data/multi_epoch_analysis_results_polynomial.txt}; 
      \addlegendentry{Average $\lambda$-Net $n\myeq4$}
      
      \addplot[color=Dandelion, style=dashed] table[x expr=\thisrowno{0} , y=n5_lambda, col sep=semicolon] {./data/multi_epoch_analysis_results_polynomial.txt}; 
      \addlegendentry{Average $\lambda$-Net $n\myeq5$}
      
      \addplot[color=PineGreen, style=dashed] table[x expr=\thisrowno{0} , y=n6_lambda, col sep=semicolon] {./data/multi_epoch_analysis_results_polynomial.txt}; 
      \addlegendentry{Average $\lambda$-Net $n\myeq6$}
      
      \addplot +[color=MidnightBlue] table[x expr=\thisrowno{0} , y=n4_int, col sep=semicolon] {./data/multi_epoch_analysis_results_polynomial.txt}; 
      \addlegendentry{$\mathcal{I}$-Net $n\myeq4$}
      
      \addplot[color=Dandelion] table[x expr=\thisrowno{0} , y=n5_int, col sep=semicolon] {./data/multi_epoch_analysis_results_polynomial.txt}; 
      \addlegendentry{$\mathcal{I}$-Net $n\myeq5$}      
      
      \addplot[color=PineGreen] table[x expr=\thisrowno{0} , y=n6_int, col sep=semicolon] {./data/multi_epoch_analysis_results_polynomial.txt}; 
      \addlegendentry{$\mathcal{I}$-Net $n\myeq6$}      
    \end{axis}

    \end{tikzpicture}
    \caption[$\lambda$ and $\mathcal{I}$-Net Error Comparison for different $\lambda$-Net Training Epochs on Cubic Polynomials]{$\lambda$ and $\mathcal{I}$-Net Error Comparison for different $\lambda$-Net Training Epochs on Cubic Polynomials}
    \label{fig:comparison_lambda_inet_poly}

\end{figure}

Within our second experiment series, we analyze the $\lambda$-Net training process and its impact on the performance of an $\mathcal{I}$-Net.  Thereby, we can show that xRAI is able to interpret neural networks very accurately at arbitrary stages of the $\lambda$-Net training process. 
Fig.~\ref{fig:comparison_lambda_inet_boolean} and  Fig.~\ref{fig:comparison_lambda_inet_poly} show the accuracy of the $\lambda$- as well as the $\mathcal{I}$-Net dependent on the number of $\lambda$-Net training epochs. Accordingly, for this experiment series we trained multiple $\mathcal{I}$-Nets, each on a dataset comprising the checkpointed weights and biases of the $\lambda$-Nets at the respective epoch. 

Starting with the results of the Boolean algebra in Fig.~\ref{fig:comparison_lambda_inet_boolean}, we can observe that astonishingly the $\mathcal{I}$-Net achieves a higher accuracy as the $\lambda$-Nets throughout all $200$ evaluated epochs and therefore is able to extract the function that should be learned by a $\lambda$-Net more accurately as the model itself has learned it. Thereby, the $\mathcal{I}$-Net performance converges very quickly compared to the $\lambda$-Nets. Especially within the first epochs, the $\mathcal{I}$-Net accuracy is significantly higher than the $\lambda$-Net accuracy which shows that the $\mathcal{I}$-Net is able to extract the target function very accurately even after a short period of $\lambda$-Net training.
Furthermore, we can observe that the overall accuracy of both, the $\lambda$- and the $\mathcal{I}$-Net slightly decreases with increasing complexity of the learning problem (i.e.~when adding more variables and therefore increasing the space of possible functions) but nevertheless, the accuracy of the $\mathcal{I}$-Net is consistently higher. However, during the first approximately $70$ training epochs, the $\lambda$-Net performance for $n\myeq6$ increases fastest followed by $n\myeq5$ and $n\myeq4$ which might seem surprising. We can explain this by the network structure of the $\lambda$ which has an increasing number of neurons for higher complexities. Therefore, the $\lambda$-Nets are able to converge faster initially.
Additionally, we can observe an accuracy decrease of $\approx 0.08$ around epoch $40$ for $n\myeq4$. We can trace this back to the fact that for $n\myeq4$ more than $50\%$ of all possible Boolean functions is used for the $\mathcal{I}$-Net training. So if we select these functions randomly it is possible that we end up with an unfavourable training set which may lead the $\mathcal{I}$-Net to overfit.

The results for polynomials (Fig.~\ref{fig:comparison_lambda_inet_poly}) are similar to the findings for the Boolean algebra described above: The $\mathcal{I}$-Net achieves a lower error than the average $\lambda$-Net and with increasing complexity of the learning problem, we have increasing error values for both models. Again, even for a low number of training epochs, we can extract high-quality information from the $\lambda$-Nets, even before they start to converge. Nevertheless, even after $200$ training epochs, when the $\lambda$-Net error just decreases slightly, the error of the $\mathcal{I}$-Net is significantly lower, even for higher complexities.
Additionally, it stands out that the $\mathcal{I}$-Net error starts to increase again slightly after about $150$ $\lambda$-Net training epochs. At approximately the same point, the $\lambda$-Net starts to converge and the error does not decrease significantly anymore. Therefore, we could assume that the $\lambda$-Net starts to overfit to the training data. This might be recognized by the $\mathcal{I}$-Net in a way that the weights do not contain enough information to extract a generalized function but the contained information is based on specific data points. However, to verify this assumption further analysis is required which is not in the scope of this paper.

In general, we showed that it is possible to extract high-quality information from already trained neural networks at arbitrary stages of the training process for different algebras. In fact, the extracted function is on average significantly closer to the actual target function compared to the $\lambda$-Nets from which the function is extracted, especially at early stages of the training process.

\subsubsection{Experiment Series 3: $\mathcal{I}$-Net Performance Comparison Different Training Data Sizes}

The third experiment observes the $\mathcal{I}$-Net performance for increasing training set sizes. The training of an $\mathcal{I}$-Net is, as already mentioned, a single-time effort. However, this single training effort can be very time consuming since a lot of $\lambda$-Nets are required which must be trained themselves. Therefore, we evaluate if and how the performance of the $\mathcal{I}$-Net can be increased once a larger set of $\lambda$-Nets is used for the training.
Fig.~\ref{fig:comparison_inet_samples_boolean} and Fig.~\ref{fig:comparison_inet_samples_poly} show the $\mathcal{I}$-Net performance for increasing sizes of the train data (i.e.~the number of $\lambda$-Nets used during the $\mathcal{I}$-Net training) along with the respective $\lambda$-Net performance which is constant since it is independent from the $\mathcal{I}$-Net training set size.

For the Boolean algebra (Fig.~\ref{fig:comparison_inet_samples_boolean}) we can observe that the $\mathcal{I}$-Net performance increases significantly when increasing the sample size. For $n\myeq4$ we can already achieve an accuracy of $1$ with $7,500$ training samples. When considering $n\myeq5$, the performance also increases significantly even for small increases in the sample size and we can achieve a accuracy of $1$ with $20,000$ training samples. If we increase the complexity even further, we are not able to achieve a perfect accuracy anymore with the maximum amount of $36,864$ samples. However, we can observe that the accuracy consistently increases when more samples are used for the training and even towards the current maximum training set size, there is still a significant performance increase. 
Accordingly, the more complex the training problem, the more beneficial and crucial is a large training data set. However, the $\mathcal{I}$-Net scales very well if we consider how the complexity increases with the number of variables. While for $n\myeq4$ there are $2^{2^4} = 65,536$ possible Boolean functions, there are $2^{2^5} = 4,294,967,296$ for $n\myeq5$ and $2^{2^6}\approx 1.8445e19$ for $n\myeq6$. Accordingly, when we consider the maximum training set size of $36,864$ there are $56.25\%$ of all possible Boolean functions considered for $n\myeq4$ while there are just $\approx 1.998590404e-13 \%$ of all possible Boolean functions considered for $n\myeq6$. Nevertheless, even for $n\myeq6$, the $\mathcal{I}$-Net achieves a higher accuracy as the respective $\lambda$-Nets already with a sample size of $2,500$.

\begin{figure}[ht]
    \begin{tikzpicture}[scale=0.90]
    \begin{axis}[
          xlabel={$\mathcal{I}$-Net Training Set Size},
          ylabel style={align=center},
          ylabel={Accuracy},
          width=1\columnwidth,
          height=0.6\columnwidth,
          line width=1.0pt,
          mark size=2.5pt,
          every axis/.append style={font=\normalsize},
          ymax=1.05,
          ymin=0.35,
          ytick={0.4, 0.6, 0.8, 1},
          xmax = 36864,
          xmin = 0,
          label style={font=\scriptsize},
          xticklabel style={
          /pgf/number format/fixed,
          /pgf/number format/precision=5,
          font=\tiny,
          },
          scaled ticks=false,
          legend columns=3,
          legend style={font=\tiny, cells={align=center}, at={(0.5, -0.3)}, anchor=north, line width=0.4pt},
          every node near coord/.style=above,
          ]
      
      \addplot +[color=MidnightBlue, style=dashed] table[x expr=\thisrowno{0} , y=n4_lambda, col sep=semicolon] {./data/samples_analysis_results_boolean.txt}; 
      \addlegendentry{Average $\lambda$-Net $n\myeq4$}
      
      \addplot[color=Dandelion, style=dashed] table[x expr=\thisrowno{0} , y=n5_lambda, col sep=semicolon] {./data/samples_analysis_results_boolean.txt}; 
      \addlegendentry{Average $\lambda$-Net $n\myeq5$}
      
      \addplot[color=PineGreen, style=dashed] table[x expr=\thisrowno{0} , y=n6_lambda, col sep=semicolon] {./data/samples_analysis_results_boolean.txt}; 
      \addlegendentry{Average $\lambda$-Net $n\myeq6$}
      
      \addplot +[color=MidnightBlue] table[x expr=\thisrowno{0} , y=n4_int, col sep=semicolon] {./data/samples_analysis_results_boolean.txt}; 
      \addlegendentry{$\mathcal{I}$-Net $n\myeq4$}
      
      \addplot[color=Dandelion] table[x expr=\thisrowno{0} , y=n5_int, col sep=semicolon] {./data/samples_analysis_results_boolean.txt}; 
      \addlegendentry{$\mathcal{I}$-Net $n\myeq5$}      
      
      \addplot[color=PineGreen] table[x expr=\thisrowno{0} , y=n6_int, col sep=semicolon] {./data/samples_analysis_results_boolean.txt}; 
      \addlegendentry{$\mathcal{I}$-Net $n\myeq6$}      
    \end{axis}

    \end{tikzpicture}
    \caption[$\mathcal{I}$-Net Accuracy Comparison for Different Training Set Sizes]{$\mathcal{I}$-Net Error Comparison for Different Training Set Sizes}
    \label{fig:comparison_inet_samples_boolean}

\end{figure}

Similar to the Boolean algebra, we can observe a significant performance for polynomials (Fig.~\ref{fig:comparison_inet_samples_poly}) when increasing the sample size for the training data. Even at a sample size of $100$ the $\mathcal{I}$-Net already achieves a similar performance as the respective $\lambda$-Nets. When increasing the sample size up to $\approx 2,500$, we can observe a significant performance, especially for higher complexities. However, even towards the maximum sample size considered in this experiment, we can observe that the error is still decreasing, especially for $n\myeq6$ which again shows that the more difficult the learning problem, the more beneficial is a larger dataset. In this case, we cannot define the complexity as previously by the number of possible functions since there is an infinite number polynomials regardless of the number of variables. Nevertheless, the function becomes more complex when more variables are added since the number of possible monomial terms increases with $n$ and therefore more complex interactions are possible. 

\begin{figure}[ht]
    \begin{tikzpicture}[scale=0.90]
    \begin{axis}[
          xlabel={$\mathcal{I}$-Net Training Set Size},
          ylabel style={align=center},
          ylabel={Mean Absolute Error},
          width=1\columnwidth,
          height=0.6\columnwidth,
          line width=1.0pt,
          mark size=2.5pt,
          every axis/.append style={font=\normalsize},
          ymax=8.5,
          ymin=0.0,
          ytick={8, 6, 4, 2, 0},
          label style={font=\scriptsize},
          xticklabel style={
          /pgf/number format/fixed,
          /pgf/number format/precision=5,
          font=\tiny,
          }, 
          scaled ticks=false,
          legend columns=3,
          legend style={font=\tiny, cells={align=center}, at={(0.5, -0.3)}, anchor=north, line width=0.4pt},
          every node near coord/.style=above,
          ]
      
      \addplot +[color=MidnightBlue, style=dashed] table[x expr=\thisrowno{0} , y=n4_lambda, col sep=semicolon] {./data/samples_analysis_results_polynomial.txt}; 
      \addlegendentry{Average $\lambda$-Net $n\myeq4$}
      
      \addplot[color=Dandelion, style=dashed] table[x expr=\thisrowno{0} , y=n5_lambda, col sep=semicolon] {./data/samples_analysis_results_polynomial.txt}; 
      \addlegendentry{Average $\lambda$-Net $n\myeq5$}
      
      \addplot[color=PineGreen, style=dashed] table[x expr=\thisrowno{0} , y=n6_lambda, col sep=semicolon] {./data/samples_analysis_results_polynomial.txt}; 
      \addlegendentry{Average $\lambda$-Net $n\myeq6$}
      
      \addplot +[color=MidnightBlue] table[x expr=\thisrowno{0} , y=n4_int, col sep=semicolon] {./data/samples_analysis_results_polynomial.txt}; 
      \addlegendentry{$\mathcal{I}$-Net $n\myeq4$}
      
      \addplot[color=Dandelion] table[x expr=\thisrowno{0} , y=n5_int, col sep=semicolon] {./data/samples_analysis_results_polynomial.txt}; 
      \addlegendentry{$\mathcal{I}$-Net $n\myeq5$}      
      
      \addplot[color=PineGreen] table[x expr=\thisrowno{0} , y=n6_int, col sep=semicolon] {./data/samples_analysis_results_polynomial.txt}; 
      \addlegendentry{$\mathcal{I}$-Net $n\myeq6$}      
    \end{axis}

    \end{tikzpicture}
    \caption[$\mathcal{I}$-Net Error Comparison for Different Training Set Sizes]{$\mathcal{I}$-Net Error Comparison for Different Training Set Sizes}
    \label{fig:comparison_inet_samples_poly}
\end{figure}

In general, we were able to show that even with a small sample size and therefore little training effort, we are able to achieve reasonable results with the $\mathcal{I}$-Net. Furthermore, we showed that when increasing the training effort, it is very likely to significantly improve the performance of the $\mathcal{I}$-Net, especially when increasing the complexity.

\section{Related Approaches}

The interpretability of blackbox models is a upcoming topic in the AI community and the terminology and classification of approaches in this field is emerging dynamically \cite{art:lipton2019} \cite{carvalho2019machine} \cite{molnar2020}.
Existing approaches regarding post-hoc interpretability generating an intrinsically interpretable model, as for instance metamodeling techniques like \textit{symbolic metamodeling}~\cite{alaa2019demystifying} or \textit{RBF-HDMR}~\cite{rbf-hdmr_shan2010metamodeling,shan2011turning}, try to fit a surrogate model which offers a higher level of explainability than the black-box model. This surrogate model can for instance be expressed by symbolic expressions comprising familiar and human-understandable, mathematical functions~\cite{alaa2019demystifying}. However, the surrogate model is usually generated based on query access to the black-box model and without considering the model internals. Therefore, their accuracy is based on proper querying. Accordingly, information that is not queried explicitly cannot be contained in the surrogate model which is not the case when the model is build based on the model internals.\todo{Findet ihr es sinnvoll diesen Vorteil hier auch darzustellen bzw ist verständlich was ich damit meine?} 
Furthermore,  Methods like \textit{NIT}~\cite{nit_tsang2018neural} are also able to generate an intrinsically interpretable model. Compared to metamodeling techniques, they also consider the model internals but thereby focus on statistical interactions inside the model rather than the overall model behaviour. Furthermore, they need to be applied ante-hoc and therefore it is not possible to use them for already trained black-box models which is often desired.
Additionally, there is a variety of methods like DeepRED \cite{deepred}, REFNE \cite{rules1_zhou2003extracting} or the Re-RX algorithm \cite{setiono2008recursive} as well as the early work of \cite{rules5_garcez2001symbolic} and \cite{rules6_taha1999symbolic} trying to convert black-box models to symbolic expressions in terms of for instance a set of rules.\todo{Habe ich mal noch hinzugefügt für ein paar referenzen. Die Paper sind teilweise recht alt, wurden uns aber auch vom IJCAI reviewer vorgeschlagen.}
The majority of interpretability approaches, including those mentioned here, focus on understanding what the black-box model has actually learned and therefore allows us to draw conclusions on the model behaviour. On the opposite, we want to uncover the target function of the black-box model and therefore are able to identify possible intentions of the model based on the training data.

\section{Conclusions and Future Work}
In this paper, we have introduced xRAI - a machine learning approach for the extraction of explicit representations of mathematical functions from trained neural networks. The presented method relies on an offline training of so-called interpretation networks ($\mathcal{I}$-Nets) for a certain family of functions. These $\mathcal{I}$-Nets enable the translation of neural network parameters (i.e.~the weights and biases) to corresponding, human understandable terms in a well-defined algebraic language. Accordingly, we are able to increase the interpretability of neural networks post-hoc by identifying and extracting the respective target function as an intrinsically interpretable function. In our evaluation we could prove that our $\mathcal{I}$-Nets are able to learn effectively the determination of polynomial or Boolean expressions based on the weights and biases of neural nets in principle. In this paper, we have focused on lower order polynomials and Boolean functions with a limited number of variables. Scaling up the approach to higher order polynomials and Boolean Functions with many Variables is subject to future work. So is the transfer of xRAI to cover further families of functions.

Furthermore, within our experiments we assume a exact algebra selection which means that the target function can always be expressed error-free using a function from the selected algebra. These assumptions allow the $\mathcal{I}$-Net to always find a well suited function, while the $\lambda$-Nets do not have this additional information during their training. Therefore, in future work, it has to be evaluated in which way the findings of this paper hold in a real-world setting since the $\mathcal{I}$-Net can always be just as good as the algebra estimation.

\bibliographystyle{named}
\bibliography{polynoms}

\end{document}